# Adaptive Traffic Signal's Safety and Efficiency Improvement by Multi-Objective Deep Reinforcement Learning Approach

**Shahin Mirbakhsh[1], Mahdi Azizi[2]**

[1,2]Department of Civil and Environmental Engineering, Shahid Chamran University of Ahvaz.

**ABSTRACT:** This research introduces an innovative method for adaptive traffic signal control (ATSC) through the utilization of multi-objective deep reinforcement learning (DRL) techniques. The proposed approach aims to enhance control strategies at intersections while simultaneously addressing safety, efficiency, and decarbonization objectives. Traditional ATSC methods typically prioritize traffic efficiency and often struggle to adapt to real-time dynamic traffic conditions. To address these challenges, the study suggests a DRL-based ATSC algorithm that incorporates the Dueling Double Deep Q Network (D3QN) framework. The performance of this algorithm is assessed using a simulated intersection in Changsha, China. Notably, the proposed ATSC algorithm surpasses both traditional ATSC and ATSC algorithms focused solely on efficiency optimization by achieving over a 16% reduction in traffic conflicts and a 4% decrease in carbon emissions. Regarding traffic efficiency, waiting time is reduced by 18% compared to traditional ATSC, albeit showing a slight increase (0.64%) compared to the DRL-based ATSC algorithm integrating the D3QN framework. This marginal increase suggests a trade-off between efficiency and other objectives like safety and decarbonization. Additionally, the proposed approach demonstrates superior performance, particularly in scenarios with high traffic demand, across all three objectives. These findings contribute to advancing traffic control systems by offering a practical and effective solution for optimizing signal control strategies in real-world traffic situations.

## INTRODUCTION

The rise in vehicle numbers on roads has led to heightened congestion, safety concerns, and negative environmental impacts (Zhu et al., 2019). Traffic signal control (TSC) has emerged as a critical strategy to address these challenges by improving road traffic safety, mitigating urban traffic congestion, and reducing vehicle emissions (Zhao and Tian, 2012; Mirbakhsh, 2022). Historically, the dominant method for traffic signal control has been fixed-time control using the Webster method. This method relies on historical traffic data to pre-determine signal timings, which remain constant throughout the day regardless of actual traffic conditions (Muralidharan et al., 2015). Actuated control systems offer more flexibility and responsiveness compared to fixed-time control systems, adjusting signal timings based on real-time vehicle presence or absence (Wang et al., 2018). However, actuated control still depends on predefined signal timing plans and may not dynamically adapt to changing traffic conditions or optimize timings based on specific objectives. Adaptive traffic signal control (ATSC) systems have emerged as a more effective alternative, adjusting signal timing in real-time to adapt to dynamic traffic demand (Stevanovic et al., 2009). Traditional ATSC systems like SCATS and SCOOT have been effective in managing traffic congestion and improving flow in urban environments (Zhao and Tian, 2012). However, their reliance on preset timings and simplified traffic models may limit adaptability and effectiveness in handling real-world traffic scenarios (McKenney and White, 2013; Shelby, 2004; Wongpiromsarn et al., 2012).

To overcome these limitations and improve the adaptability of ATSC across various scenarios, a deep reinforcement learning (DRL) approach is proposed. Unlike traditional methods, DRL doesn't rely on prior knowledge of specific traffic scenarios. Instead, it learns the best course of action through continuous interaction with the traffic environment. This has led to the widespread adoption of DRL-based approaches in developing traffic control systems with different objectives and purposes (El-Tantawy et al., 2014; Younes and Boukerche, 2015) (Arulkumaran et al., 2017; François-Lavet et al., 2018). DRL-based ATSC systems have primarily aimed to address three key objectives in traffic management: improving traffic efficiency and/or reducing carbon emissions (Ceylan and Bell, 2004; Zheng et al., 2010; Mohebifard and Hajbabaie, 2019), and enhancing safety (Stevanovic et al., 2013; Stevanovic et al., 2015). While these systems have shown promising results in achieving their respective goals, inconsistent findings regarding the impact of specific-oriented ATSC systems on other traffic management objectives have been reported. Some studies indicate that mobility-oriented ATSC schemes can negatively affect driver behavior and increase traffic conflicts, while others suggest they can enhance traffic efficiency and reduce conflicts (Fink et al., 2016; Tageldin et al., 2014). These conflicting findings underscore the complexity of the relationship between different traffic management indicators and emphasize the need for a comprehensive and balanced approach in ATSC system design. Gong et al. (2020) standout as one of the few studies that successfully address both

# Adaptive Traffic Signal's Safety and Efficiency Improvement by Multi-Objective Deep Reinforcement Learning Approach

traffic safety and efficiency in their ATSC system by employing a multi-objective RL approach and integrating a real-time crash risk model into their system. The study demonstrates the proactive enhancement of traffic safety while optimizing traffic efficiency by considering both objectives simultaneously.

These findings underscore the importance of integrating safety considerations into ATSC systems to achieve a balanced approach that enhances both safety and efficiency. However, despite the growing recognition of the significance of environmental sustainability, there exists a notable research gap in integrating environmental sustainability with safety and efficiency in the domain of traffic signal control. Addressing environmental sustainability alongside safety and efficiency is crucial for formulating comprehensive and sustainable traffic control strategies. This study aims to bridge the aforementioned research gaps by introducing an efficient Adaptive Traffic Signal Control (ATSC) scheme that proactively improves traffic safety, efficiency, and decarbonization simultaneously, leveraging a Deep Reinforcement Learning (DRL) approach. To assess the performance of the proposed TSC system in attaining these three objectives, specific indicators such as cumulative traffic conflicts, waiting time, and vehicle carbon emissions are utilized. These indicators are integrated while considering trade-offs among safety, efficiency, and carbon emissions. The initial weights assigned to these indicators can be determined through expert opinions, public sentiments, or governmental policies. Furthermore, these weight values are open to updates, enabling adaptation to specific circumstances, all aimed at ensuring optimal outcomes. These indicators form the basis for comparing the performance of our proposed TSC system against well-established benchmark TSC algorithms. The evaluation and comparison are conducted at a simulated single intersection modeled based on real-world traffic flow conditions in Changsha, Hunan, China. This study contributes to traffic control systems research by proposing a novel multi-objective DRL-based adaptive traffic signal control (ATSC) algorithm. By integrating DRL techniques and considering multiple objectives, including safety, efficiency, and decarbonization, the proposed algorithm offers a fresh approach to optimizing traffic signal control in a more comprehensive and effective manner. The study enhances existing knowledge in traffic control system research and provides valuable insights for the development of advanced ATSC algorithms with enhanced performance and capabilities.

## 2. RELATED WORK
### 2.1. Efficiency-oriented Adaptive Traffic Signal Control (ATSC)
The majority of existing research on traffic signal control has focused primarily on optimizing signal timing to enhance traffic efficiency, aiming to minimize waiting time, reduce delays, and improve the overall flow of vehicles through signalized intersections. These studies typically utilize mathematical modeling or deep reinforcement learning (DRL) techniques tailored to specific intersection scenarios (Ceylan and Bell, 2004; Zheng et al., 2010; Mohebifard and Hajbabaie, 2019).

For example, Kumar et al. (2020) introduced a Dynamic and Intelligent Traffic Light Control System (DITLCS) designed to address efficiency issues related to vehicles with varying priorities. DITLCS integrates deep reinforcement learning and fuzzy inference system techniques to optimize signal timings in real-time, resulting in reduced waiting time and enhanced intersection throughput. This approach effectively manages uncertain traffic situations and adjusts to dynamic traffic demands, making it well-suited for real-world traffic control applications.

Similarly, Mao et al. (2021) developed an optimization framework that combines machine learning algorithms with Genetic Algorithms (GA) to tackle efficiency issues arising from incidents. This framework surpasses the original GA algorithm and significantly diminishes total travel time during incident conditions.

### 2.2. Safety-oriented Adaptive Traffic Signal Control (ATSC)
Intersections serve as critical points where various streams of traffic converge, leading to a higher risk of conflicts and accidents. Prioritizing safety at intersections is crucial for maintaining a secure transportation system. Traffic signal control plays a pivotal role in regulating the movement of road users at intersections, and optimizing these control methods can significantly enhance safety. Numerous studies have focused on developing safety-oriented ATSC systems to mitigate crash risks or traffic conflicts.

For instance, Sabra et al. (2013) proposed a safety-oriented ATSC based on a self-developed traffic conflict prediction model. This system adjusts cycle length, splits, offsets, and left-turn phase sequence using a four-stage algorithm, effectively reducing conflicts and enhancing safety.

In connected vehicle (CV) scenarios, researchers have developed ATSC systems that utilize real-time safety evaluation models to optimize safety. Reyad et al. (2021) and Essa and Sayed (2020) implemented ATSC systems integrating CV data and employing real-time safety evaluation models to optimize signal timings and improve safety at intersections. By leveraging information from connected vehicles, these systems proactively adjust signal timings to mitigate potential safety hazards.

Furthermore, Ghoul and Sayed (2021) proposed a signal-vehicle coupled control system that incorporates ATSC and speed advisories for optimizing real-time safety. This system utilizes CV data and integrates ATSC strategies with speed advisories to effectively manage traffic flow and ensure safer interactions between vehicles at intersections.

# Adaptive Traffic Signal's Safety and Efficiency Improvement by Multi-Objective Deep Reinforcement Learning Approach

Through the incorporation of advanced algorithms, predictive models, and real-time data, safety-oriented ATSC systems can effectively reduce conflicts, mitigate potential safety hazards, and enhance intersection safety.

**2.3. Multi-objective Adaptive Traffic Signal Control (ATSC)**

With increasing concern about the environmental impact of the transport sector, there is a growing emphasis on environmental sustainability in traffic signal control. Researchers highlight the importance of addressing this impact and developing strategies, such as ATSC, to promote sustainability. However, it's notable that the environmental sustainability aspect is often not explicitly considered as a separate goal of ATSC systems. Instead, ATSC systems primarily focus on improving traffic efficiency, reducing carbon emissions, and enhancing overall intersection performance.

For instance, Tan et al. (2017) proposed a bi-objective programming model to minimize overall delay and emission increment by jointly determining optimal signal timing and speed limit. Mirbakhsh et al in 2023 descripted CAV system as a control on traffic and chaos. Boukerche et al. (2021) introduced an ATSC scheme using DRL techniques with efficiency optimization rewards and speed modules to smooth traffic flow, achieving both efficiency improvement and decarbonization. Kou et al. (2018) developed a multi-objective optimization model considering delay, stops, and emissions, utilizing Genetic Algorithms to calculate optimal signal timing. Additionally, the eco-approach and departure (EAD) application of connected vehicles and smart antenna technology have been employed in ATSC systems, improving traffic efficiency and reducing emissions simultaneously (Hao et al., 2018; Joyo et al., 2020).

However, the consideration of both traffic efficiency and safety in ATSC systems has been relatively limited. Gong et al. (2020) made a significant contribution to the field by optimizing traffic safety and efficiency simultaneously in an ATSC system. They integrated a crash risk model into the RL framework to guide the learning process and prioritize safety. Employing a weighted sum approach, they balanced the trade-off between traffic safety and efficiency. Evaluation results demonstrated the effectiveness of their approach in improving both traffic efficiency, measured by reducing the cumulative waiting time of queued vehicles, and traffic safety, measured by reducing crash risks. However, compared with an adaptive traffic signal optimizing only traffic efficiency, the proposed algorithm resulted in a slight deterioration of traffic efficiency while significantly enhancing traffic safety. This suggests a potential trade-off between traffic safety and efficiency, where improvements in one objective might come at the expense of the other.

Overall, there have been limited published studies explicitly addressing the simultaneous optimization of all three objectives of traffic efficiency, safety, and decarbonization in ATSC systems. Most research in this area has primarily focused on optimizing one or two objectives while giving less attention to the integration of all three goals. Further research and development efforts are necessary to advance the integration of all three goals and develop comprehensive and sustainable ATSC systems.

**3. THE PROPOSED ADAPTIVE TRAFFIC SIGNAL CONTROL**

The main objective of this research is to improve safety, streamline traffic flow, and reduce carbon emissions at intersections using a proficient Adaptive Traffic Signal Control (ATSC) system based on the Dueling Double Deep Q Network (D3QN), referred to as ATSC-SED. This section delves into the research challenges, underlying assumptions, and provides an in-depth explanation of the proposed ATSC algorithm.

**3.1. Research problem statement**

In this setting, a signalized intersection connects to four access roads, serving as an illustration of pertinent Traffic Signal Control (TSC) definitions and concepts. The intersection comprises four entry approaches, each accommodating two-way traffic with six lanes, facilitating travel in distinct directions (North/West/South/East, abbreviated as N, W, S, E). Figure 1(a) depicts the lane layout for vehicles entering the intersection from all directions, along with the regulations governing vehicular movements in each lane. Typically, the rightmost lane accommodates vehicles turning right or proceeding straight, the leftmost lane facilitates left turns, and the middle lane caters to vehicles proceeding straight. Although intersection layouts may differ across regions, these principles are adaptable to various intersection configurations, with adjustments tailored to traffic demand and intersection geometry (e.g., T-junctions).

# Adaptive Traffic Signal's Safety and Efficiency Improvement by Multi-Objective Deep Reinforcement Learning Approach

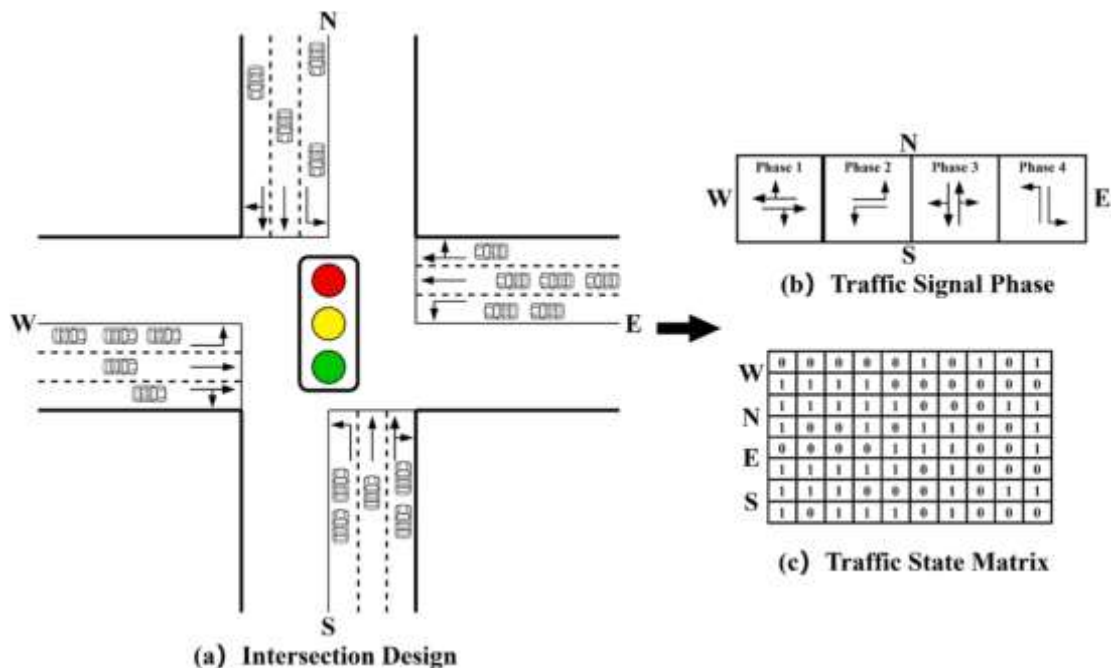

Fig 1. Intersection Configuration

A traffic signal phase encompasses the timing of red, yellow, and green signals in entry lanes, facilitating the passage of vehicles through the intersection area while minimizing conflicts. Figure 1(b) showcases four traffic signal phases devised to prevent conflicting movements and safeguard left-turn maneuvers. Phases 1 and 3 designate green signals for vehicles in the middle and rightmost lanes to proceed straight or turn right in the W-E and N-S directions, respectively. Conversely, Phases 2 and 4 allocate green signals for vehicles in the leftmost lane to execute left turns in the W-E and N-S directions, respectively. During the activation of a particular phase for corresponding lanes, signals for other lanes transition to red.

### 3.2. Real-time safety evaluation

Previous studies have utilized real-time crash risk models, such as extreme value modeling, to assess the relationship between future crash potential (a traffic safety measure) and traffic conditions associated with signal timing. While these models offer high accuracy in crash risk assessment, they rely on police-reported crash data, which may be flawed due to underreporting, subjectivism, and information omission. Additionally, establishing these models requires waiting for sufficient crashes to occur in specific traffic scenarios, presenting a moral dilemma for researchers.

Alternatively, surrogate safety measures (SSMs) provide a quick and reliable way to assess traffic safety without relying on crash data. SSMs are typically calculated based on microscopic traffic parameters to characterize safety risks. They can be categorized into time metrics, distance metrics, and deceleration metrics, such as time-to-collision (TTC), stopping headway distance (SHD), and deceleration rate to avoid a crash (DRAC). At intersections, potential crash risk often arises from abrupt changes in vehicle acceleration and deceleration due to signal adjustments, leading to rear-end conflicts. Therefore, Time-to-Collision (TTC) is a suitable indicator for assessing traffic safety at intersections, establishing the relationship between signal timing and rear-end conflicts.

The TTC is calculated using the formula:

$$TTC(s) = \frac{d}{v_F - v_L}$$

where $d$ is the distance between the vehicle and its leading vehicle, and $v_F$ and $v_L$ are the speeds of the vehicle and its leading vehicle, respectively. A traffic conflict or crash risk exists when TTC(s) is less than three seconds; otherwise, it does not exist.

### 3.3. Carbon emission evaluation

Traffic control strategies and signal timing can affect traffic flow at intersections, indirectly influencing the carbon dioxide emissions of vehicles. Optimized signal timing strategies aim to reduce vehicle waiting time and the number of stops, thereby decreasing emissions at intersections. Traffic models and simulation tools can assess the effects of different signal control strategies on emissions.

Carbon dioxide emissions are evaluated using the pollutant emission model of SUMO (Simulation of Urban MObility), which defines emission quantity (g/h) as a function of vehicular current engine power using typical emission curves over power (CEPs). The total carbon dioxide emissions $PE$ are defined as:

# Adaptive Traffic Signal's Safety and Efficiency Improvement by Multi-Objective Deep Reinforcement Learning Approach

$$PE = (P_{Roll} + P_{Air} + P_{Accel} + P_{Grad})\eta_{gearbox}$$

where:

$P_{Roll}$: Rolling resistance power,
$P_{Air}$: Air resistance power,
$P_{Accel}$: Power to accelerate,
$P_{Grad}$: Power to overcome gradients,
$\eta_{gearbox}$: Efficiency of the gearbox.

The components are further defined as per specific parameters such as vehicular masses, load masses, gradients, air density, and drag coefficients, among others.

### 3.5 The deep reinforcement learning (DRL) framework

for traffic signal control (TSC) is designed to address the traffic control problem by formulating it into a DRL setting. This framework consists of three main components: the intersection environment, the traffic signal controller (agent), and the signal control components.

1. **Intersection Environment**: The traffic condition at the intersection is represented as the state ($s$). This state captures relevant information about the current traffic flow, such as vehicle density, speed, and occupancy at various lanes and approaches.
2. **Traffic Signal Controller (Agent)**: The traffic signal controller acts as the agent in the DRL framework. It observes the state ($s$) of the intersection and selects an action ($a$), which corresponds to choosing the appropriate traffic signal phase. The agent's decision-making process is guided by a deep reinforcement learning algorithm, specifically the Double Deep Q Network (D3QN). The action chosen by the agent influences the traffic flow at the intersection.
3. **Signal Control Components**: These components execute the selected action ($a$) determined by the traffic signal controller. They are responsible for implementing the chosen traffic signal phase, either maintaining the current phase or transitioning to a new one. The signal control components interact with the physical infrastructure of the intersection to adjust traffic signal timings accordingly.

The interaction between these components is depicted in Figure 2 of the framework. At each time step $t$, the agent observes the current state ($S_t$) of the intersection environment and calculates the action ($a_t$) based on this state using the D3QN algorithm. The reward ($r_t$) for the chosen action is determined by a reward function that combines multiple objectives, including traffic safety, efficiency, and decarbonization.

The signal control components then execute the chosen action, adjusting the traffic signal phase accordingly. The resulting change in the traffic condition leads to a new state ($S_{t+1}$), and the process repeats for the next time step. The experiences of the agent, consisting of state-action-reward-state tuples ($S_t, a_t, r_t, S_{t+1}$), are stored in memory and used for experience replay, allowing the agent to learn and update its policy over time.

This framework enables the traffic signal controller to learn optimal signal control policies through interaction with the intersection environment, ultimately leading to improved traffic safety, efficiency, and decarbonization.

### 3.6 Agent Configuration

The agent design in deep reinforcement learning (DRL)-based Adaptive Traffic Signal Control (ATSC) systems is a crucial aspect that determines how the agent interacts with the traffic environment and learns to make optimal decisions regarding traffic signal control. Here are some key considerations in agent design:

1. State Representation: The state representation encapsulates information about the current traffic environment. This can include factors such as vehicle queue length, vehicle speed, occupancy at various lanes, and other relevant traffic parameters. Discrete Traffic State Encode (DTSE) is one method used to encode traffic state information into a format suitable for input into the DRL algorithm.
2. Action Space: The action space defines the possible actions that the agent can take in response to the observed state. In the context of ATSC, actions typically involve selecting a specific Traffic Signal Phase (TSP), maintaining the current phase, or transitioning to the next phase. Additionally, the agent may adjust the duration of the selected phase based on the traffic conditions.
3. Reward Function: The reward function evaluates the desirability of the agent's actions based on their impact on traffic flow and system objectives. Rewards are scalar values that provide feedback to the agent about the quality of its decisions. Common metrics used in the reward function include vehicle waiting time, queue length, vehicle delay, and other relevant performance indicators.
4. Learning Algorithm: The learning algorithm governs how the agent updates its policy based on observed states, actions, and rewards. Deep Q-Networks (DQN) and its variants, such as Double DQN (DDQN) and Dueling DQN, are commonly used in ATSC systems to approximate the action-value function and learn optimal policies.

# Adaptive Traffic Signal's Safety and Efficiency Improvement by Multi-Objective Deep Reinforcement Learning Approach

By carefully designing the agent to incorporate relevant traffic information, define appropriate actions, and construct an effective reward function, DRL-based ATSC systems can learn to make intelligent decisions that optimize traffic flow, enhance efficiency, and improve overall system performance.

### 3.6.1 State

complex and dynamic environment, requiring a comprehensive representation of the traffic state to capture the nuances of vehicle movement and interaction. In this study, the traffic state is designed using a method based on Discrete Traffic State Encode (DTSE), which offers advantages over traditional approaches like vehicle queue length or traffic flow.

1. Non-uniform quantization: Traditional methods often divide the intersection into uniform segments, which may not effectively capture the varying density of traffic across different areas of the intersection. Non-uniform quantization allows for more flexibility in defining the boundaries of each cell based on the actual traffic distribution.
2. One-hot encoding: Each cell resulting from the non-uniform quantization is represented using one-hot encoding. This encoding scheme assigns a binary vector to each cell, where only one element is active (set to 1) to indicate the presence of vehicles in that cell, while all other elements are inactive (set to 0).

By employing DTSE with non-uniform quantization and one-hot encoding, the traffic state representation becomes more expressive, capturing both the spatial distribution of vehicles and their movement dynamics within the intersection. This detailed representation enhances the agent's ability to perceive and react to the complex traffic conditions effectively.

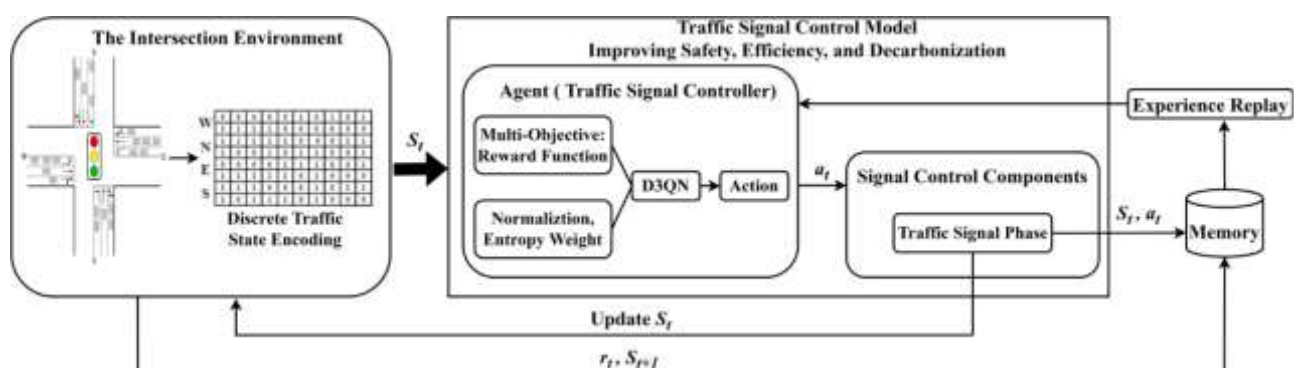

**Fig 2. Objective framework**

The approach of representing the traffic state through non-uniform quantization and one-hot encoding based on DTSE offers a simplified yet effective way to capture the main traffic characteristics near the intersection. By dividing each lane into cells and encoding the presence of vehicles within these cells using a binary representation, the proposed method provides a compact yet informative representation of the traffic state.

Compared to methods such as real-time image representation or uniform lane division, this approach simplifies the traffic information while still retaining essential details about vehicle positions and movements near the intersection. The use of a structured grid of cells allows for a systematic representation of the traffic state, enabling the agent to quickly understand the current traffic conditions and make informed decisions.

With this simplified representation, the agent can efficiently process the traffic state information and accelerate the convergence of the model. By focusing on the main traffic characteristics near the intersection, the proposed method strikes a balance between complexity and effectiveness, facilitating robust and efficient traffic signal control.

### 3.6.2 Action

The action space is defined to encompass four distinct traffic signal phases, each corresponding to a specific configuration of green signal allocation for vehicles at the intersection. These signal phases are denoted as WEG, WELG, NSG, and NSLG, representing different movements of vehicles in the W-E and N-S directions.

- WEG signifies the green signal for vehicles to proceed straight and turn right in the W-E direction.
- WELG indicates the green signal for vehicles to turn left in the W-E direction.
- NSG represents the green signal for vehicles to proceed straight and turn right in the N-S direction.
- NSLG denotes the green signal for vehicles to turn left in the N-S direction.

The design ensures rationality and standardization of traffic signal operations by setting a minimum duration of 10 seconds for each traffic signal phase. Additionally, a four-second yellow signal is implemented between two adjacent traffic signal phases to provide a transition period for traffic flow adjustments. This structured action space enables the agent to make informed decisions regarding

# Adaptive Traffic Signal's Safety and Efficiency Improvement by Multi-Objective Deep Reinforcement Learning Approach

the allocation of green signals based on the current traffic conditions, thereby optimizing the intersection's performance and achieving higher rewards.

### 3.6.2 Reward

The reward function in the DRL-based ATSC framework is crucial for guiding the agent to learn an optimal policy that achieves the desired objectives. In this study, three rewards are designed to evaluate actions based on traffic safety, efficiency, and decarbonization.

1. Traffic Safety Reward ($r(t)CTC$): The goal of this reward is to minimize traffic conflicts and reduce crash risk. Cumulative Traffic Conflict (CTC) is used as the indicator of traffic safety, calculated based on the Time-to-Collision (TTC) metric. Specifically, CTC represents the cumulative number of instances where the TTC value of vehicles on the road is less than three seconds, indicating potential conflicts. The reward for traffic safety is designed as the difference in CTC between the current and previous actions. If the cumulative traffic conflict decreases due to the change in traffic signal phase (action at), the agent receives a positive reward; otherwise, it is penalized.

$r(t) CTC = -CTC(t+1) - CTC(t)$

2. Traffic Efficiency Reward ($r(t)CWT$): This reward aims to minimize vehicle waiting times at the intersection, thereby enhancing traffic flow and reducing delays. Cumulative Waiting Time (CWT) is used as the indicator of traffic efficiency, representing the total waiting time of vehicles stopped at the intersection area. The reward for traffic efficiency is calculated as the difference in CWT between the current and previous actions. Similar to the traffic safety reward, if the cumulative waiting time decreases due to the change in traffic signal phase, the agent receives a positive reward; otherwise, it is penalized.

$r(t) CWT =- CWT(t+1) - CWT(t)$

3. Decarbonization Reward ($r(t)CDE$): The objective of this reward is to minimize carbon dioxide emissions (CDE) from vehicle operations, thereby promoting environmental sustainability. Cumulative Carbon Dioxide Emissions (CDE) is used as the indicator of decarbonization, representing the total amount of carbon dioxide emitted by vehicles. The reward for decarbonization is calculated as the difference in CDE between the current and previous actions. If the cumulative carbon dioxide emissions decrease due to the change in traffic signal phase, the agent receives a positive reward; otherwise, it is penalized.

$r(t) CDE = - PE(t+1) - PE(t)$

These reward functions provide a comprehensive evaluation of the agent's actions, considering the key objectives of improving traffic safety, efficiency, and environmental sustainability at the intersection. By optimizing these rewards, the agent can learn an effective policy for traffic signal control that balances these objectives and enhances overall intersection performance.

## 4. PERFORMANCE EVALUATION

This section outlines the experimental parameters of the simulation scenario and algorithm, detailing the compared Traffic Signal Control (TSC) methods and evaluation metrics. The algorithm for Adaptive Traffic Signal Control with Signal Event Detection (ATSC-SED) is developed, trained, and tested within a simulated single intersection environment constructed using Simulation of Urban Mobility (SUMO, Version 1.14.1). The performance evaluation of ATSC-SED takes place within this simulated intersection, utilizing real-world traffic flow data obtained from photography recordings.

To assess its efficacy, ATSC-SED is compared against both traditional real-world TSC methods and a Deep Reinforcement Learning (DRL)-based TSC scheme, which focuses solely on optimizing traffic efficiency. This comparison serves to gauge the effectiveness and efficiency of ATSC-SED in managing traffic flow within the intersection.

### 4.1 Simulation Environment

The simulation scenario faithfully replicates a classical signalized intersection, specifically a right-angled, four-legged junction situated at the intersection of Liangtang Road and Shizhong Road in Changsha City. This selection is intentional, chosen for its proximity to community buildings and a middle school. Data collection during the morning peak period offers an excellent environment for studying scenarios marked by uneven traffic flow, high volume, and dynamic traffic conditions. These conditions necessitate the strategic scheduling of vehicle right-of-way during peak demand situations.

As depicted in Figure 5(a), this intersection comprises four entering approaches, each consisting of two-way six lanes, each with a length of 500 meters. Within this setup, vehicles adhere to standard traffic rules: they turn left from the leftmost lane, proceed straight in the middle or rightmost lane, and make right turns from the rightmost lane (as illustrated in Figure 3). Importantly, in many countries, vehicles in the rightmost lane are permitted to make right turns at a red signal without causing conflicts, adding a layer of complexity to traffic management at the intersection.

**Adaptive Traffic Signal's Safety and Efficiency Improvement by Multi-Objective Deep Reinforcement Learning Approach**

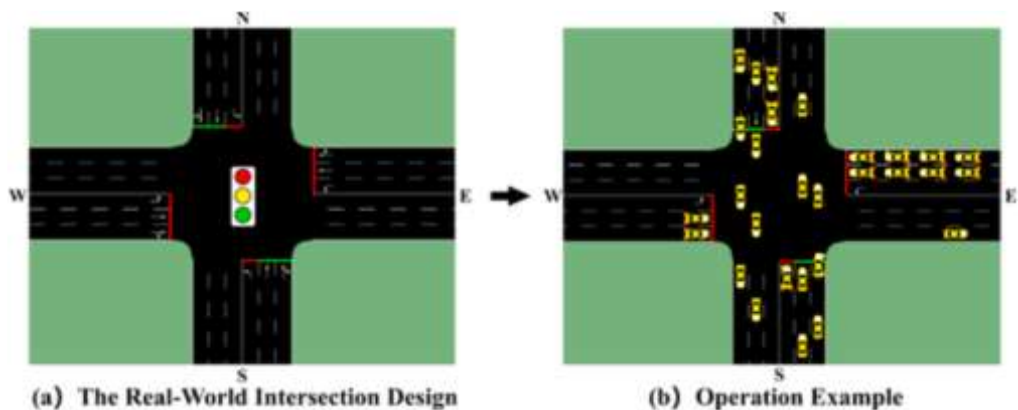

Fig 3. Realistic Signalized Intersection

**4.2 Algorithm setting**
Agent Parameters: These parameters define the characteristics and behavior of the ATSC-SED agent within the simulation environment.
TSC Model (D3QN): This denotes the specific model or algorithm used for Traffic Signal Control, in this case, D3QN, which likely refers to a variant of Deep Q-Networks tailored for traffic signal control tasks.
Memory: Refers to the memory capacity or replay buffer size used by the reinforcement learning agent to store past experiences for training purposes.
Initial Weight Assignments:
The weights assigned to safety, efficiency, and carbon emissions in the decision-making process are initially set to 0.5, 0.25, and 0.25, respectively. These weight values are chosen to align with local policies or specific requirements governing traffic management. It's emphasized that these initial weight assignments predominantly influence the agent's initial decision-making behavior, with minimal subsequent impact on weight calculations during runtime.
Importance of Initial Weights:
The initial weight assignments play a crucial role in guiding the agent's initial decision-making behavior, shaping its priorities regarding safety, efficiency, and environmental considerations. However, as the agent interacts with the environment and learns from experience, these initial weights may be dynamically adjusted based on reinforcement learning principles and the observed performance of the TSC algorithm.
By configuring the ATSC-SED algorithm with appropriate parameters and initial weight assignments, it can effectively adapt to varying traffic conditions and optimize traffic signal control decisions to achieve desired objectives, such as improving traffic flow efficiency while ensuring safety and minimizing environmental impact.
Arrival rates and vehicle counts for the selected algorithm settings are shown in Figure 4.

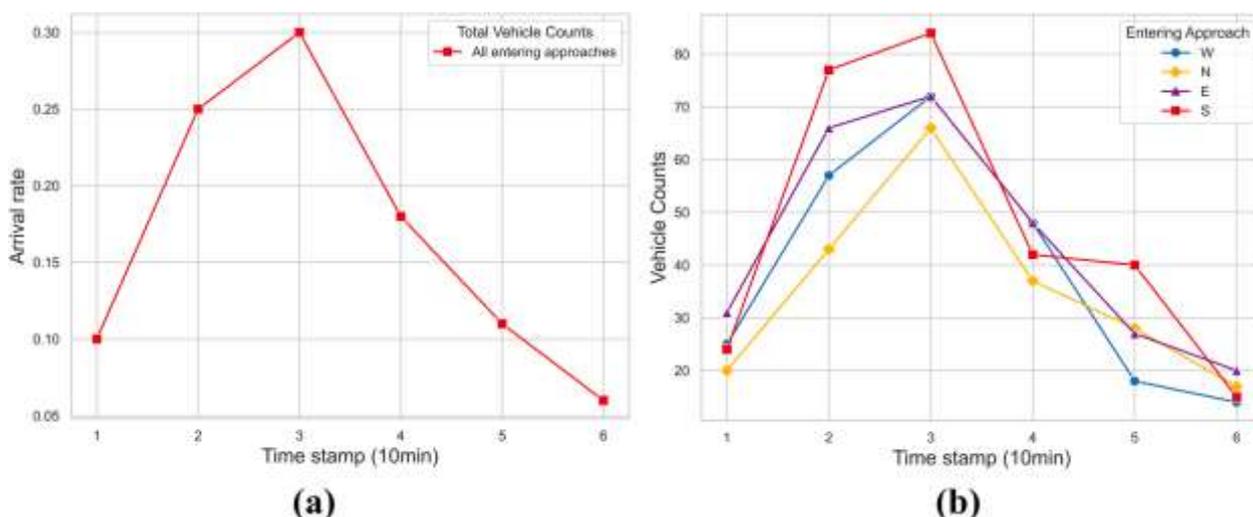

Fig 4. Vehicle Arrival versus Vehicle Count

# Adaptive Traffic Signal's Safety and Efficiency Improvement by Multi-Objective Deep Reinforcement Learning Approach

## 4.3 Results

To analyze the experimental results and evaluate the performance of various Traffic Signal Control (TSC) methods, we conducted simulations using real-world traffic flow data. The following TSC methods were implemented and compared:

1. **ATSC-SED (Adaptive Traffic Signal Control with Signal Event Detection):** The proposed scheme, which incorporates signal event detection and adaptive control mechanisms.
2. **FTC (Fixed-Time Control):** A traditional TSC method where signal timings remain constant regardless of traffic conditions.
3. **AC (Actuated Control):** Another traditional TSC method that adjusts signal timings based on detected vehicle presence or demand.
4. **ATSC-SCOOT:** An adaptive TSC method that optimizes signal timings based on real-time traffic flow data.
5. **ATSC-TE (Traffic Efficiency):** A variant of adaptive TSC focused solely on optimizing traffic efficiency without considering other factors like safety or emissions.

Separate measurements were taken for each TSC method, assessing their cumulative, average, and real-time performance across various evaluation metrics. These metrics include but are not limited to:

- **Average Delay:** The average time vehicles spend waiting at the intersection.
- **Travel Time:** The time taken by vehicles to traverse the intersection.
- **Throughput:** The rate at which vehicles successfully pass through the intersection.
- **Safety:** Metrics related to the occurrence of conflicts, collisions, or other safety-related incidents.
- **Environmental Impact:** Metrics such as carbon emissions or fuel consumption, reflecting the ecological footprint of traffic operations.

These evaluation metrics provide insights into the effectiveness, efficiency, safety, and environmental impact of each TSC method. By comparing the performance of ATSC-SED against traditional and adaptive TSC approaches, we can identify strengths, weaknesses, and areas for improvement in traffic signal control strategies. This analysis serves as a basis for discussion and further refinement of TSC methods to enhance urban traffic management and optimize intersection operations. Summary of results are shown in the figures. The converged speed for two algorithms is shown in Figure 5.

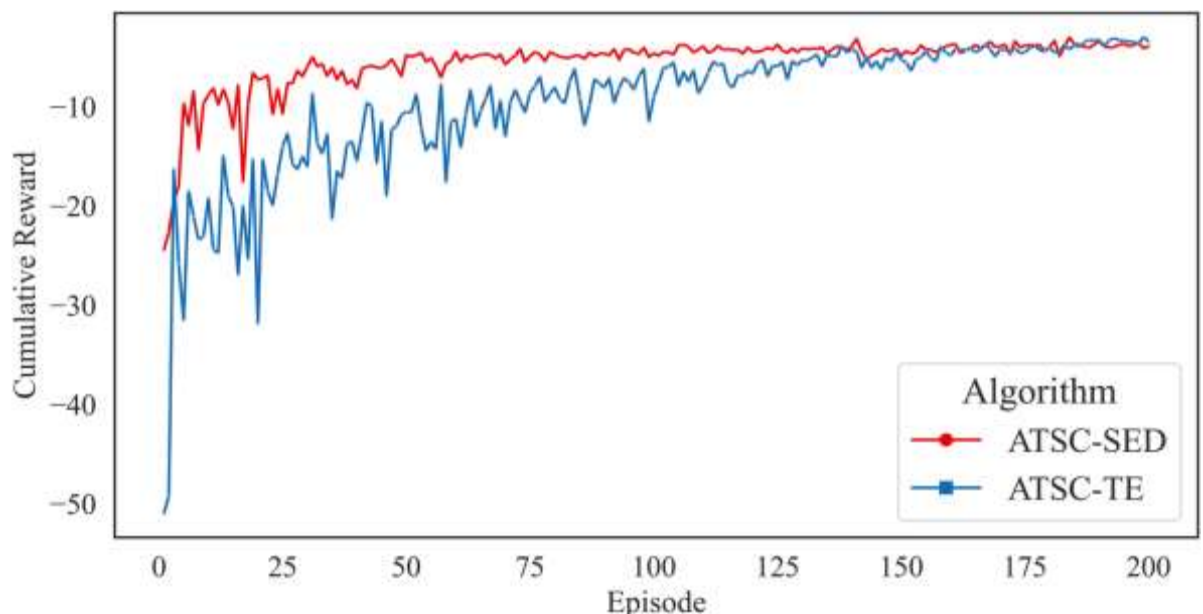

**Fig 5. The converged speed ATSC-SED and ATSC-TE at the algorithm training process.**

# Adaptive Traffic Signal's Safety and Efficiency Improvement by Multi-Objective Deep Reinforcement Learning Approach

Figures 6 and 7 show performance comparison for different measure between the two algorithms.

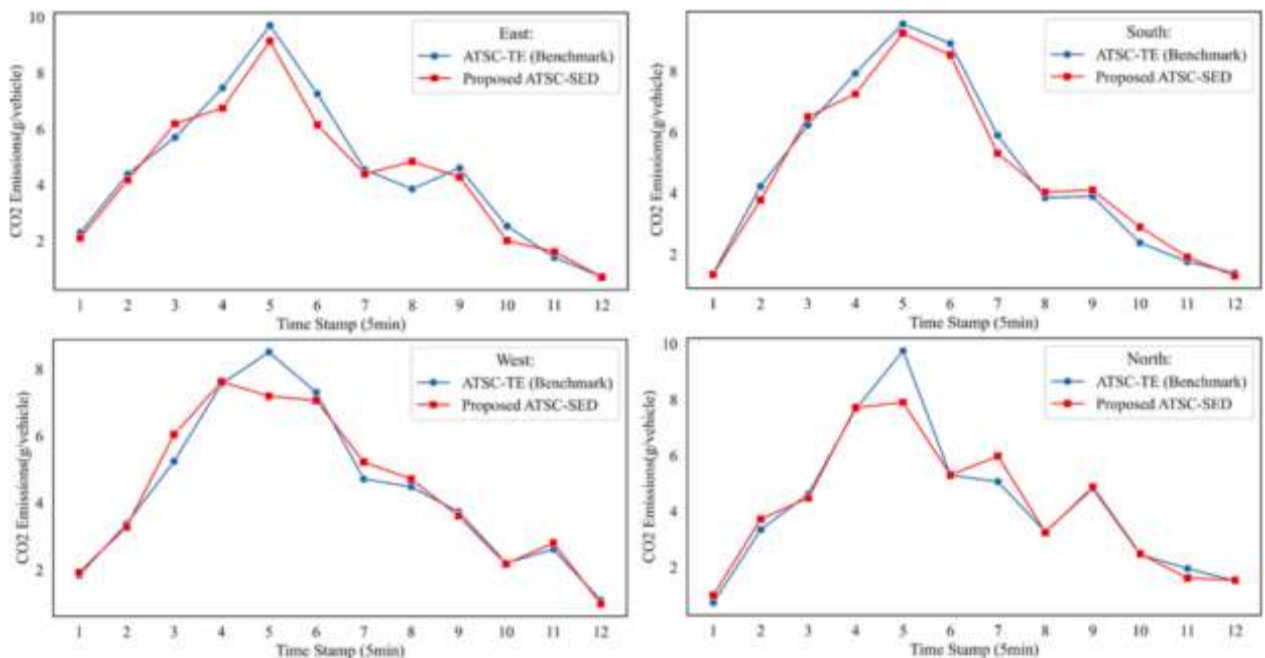

**Fig. 6. Real-time performance comparison on traffic conflict rates between ATSC-SED and ATSC-TE.**

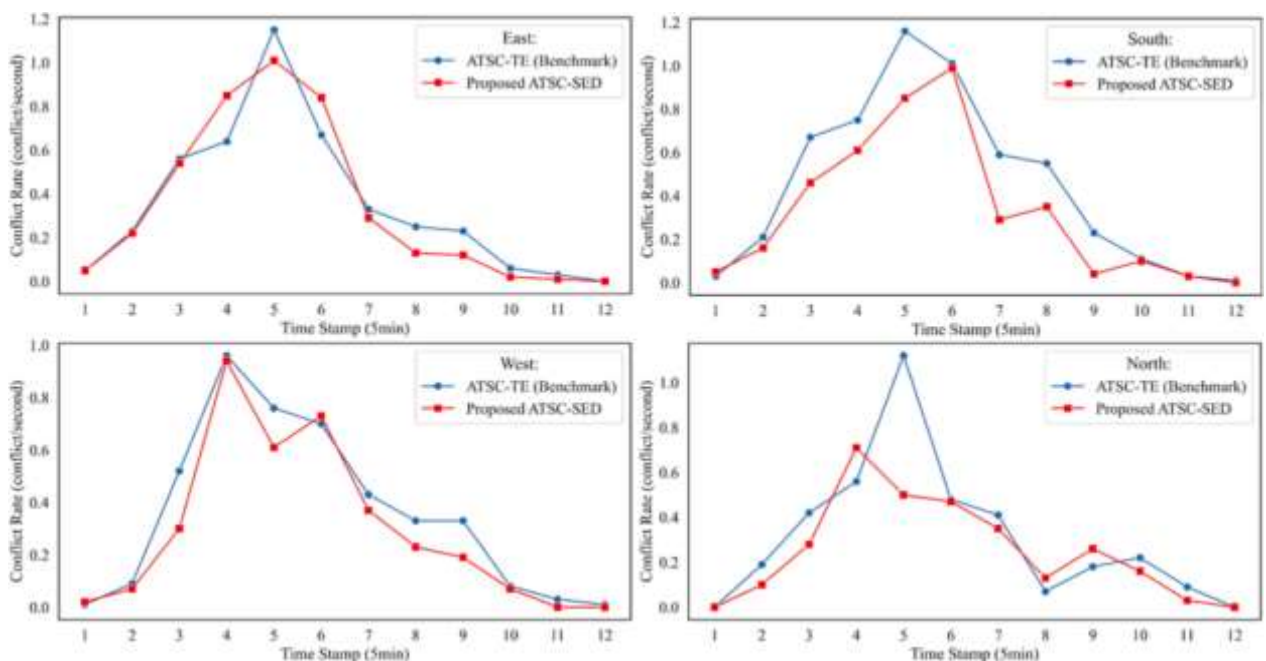

**Fig. 7. Cumulative waiting time on each approach between ATSC-SED and ATSC-TE.**

## 6. CONCLUSION

To concurrently improve traffic safety, efficiency, and environmental sustainability, the paper introduces an adaptive traffic control algorithm that optimizes these objectives in a coordinated manner. The algorithm utilizes high-resolution real-time traffic data to select appropriate signal phases every second, aiming to reduce waiting time, conflicts, and vehicle carbon emissions. To tackle the discrete optimization problem, a multi-objective deep reinforcement learning framework employing Dueling Double Deep Q Network (D3QN) is employed. The dimensions of safety, efficiency, and decarbonization objectives are unified using a normalized model and entropy weight method to construct an integrated synthetic reward function. This initiative adaptive traffic control algorithm is trained and tested in a simulated isolated intersection in Changsha, Hunan, China, using observed traffic data. The well-trained algorithm's performance is evaluated based on traffic conflict, waiting time, and carbon dioxide emissions. The results

# Adaptive Traffic Signal's Safety and Efficiency Improvement by Multi-Objective Deep Reinforcement Learning Approach

indicate an improvement in traffic efficiency, safety, and carbon emissions compared to benchmark algorithms. However, there is a slight increase in waiting time (0.6%) compared to the D3QN algorithm solely focused on efficiency optimization, suggesting a potential trade-off between safety/decarbonization objectives and efficiency. The findings underscore the benefits of a multi-objective approach in traffic signal control and stress the importance of holistic traffic management.

Further analysis of the proposed algorithms under different traffic demand scenarios reveals significant improvements and performance by DRL-based ATSC systems, particularly at higher traffic volumes. In such scenarios, the DRL-based ATSC system exhibits reductions in traffic conflicts (18.06%), waiting time (14.17%), and carbon emissions (3.51%) compared to the D3QN algorithm with efficiency optimization only. Thus, DRL-based ATSC systems are particularly advantageous in extremely busy intersections with high traffic volumes.

While offering substantial improvements in safety, efficiency, and decarbonization, the proposed ATSC system may incur higher implementation costs compared to traditional traffic control methods due to initial investment in developing and deploying the DRL-based algorithm, as well as infrastructure and hardware requirements for real-time data collection and communication. These implementation costs could pose a barrier to widespread application and adoption in real-world scenarios. Given the exceptional performance of DRL-based ATSC systems in extremely busy intersections and considering associated implementation costs, it is recommended to install these systems at critical intersections with high traffic volumes and significant safety, efficiency, and environmental concerns.The study has limitations, focusing on a specific single intersection during peak hours, and its applicability to other locations and non-peak hours is uncertain. Additionally, it primarily focuses on optimizing safety, efficiency, and decarbonization without establishing specific uniform standards or conducting detailed cost-benefit analyses. Future research could explore monetary valuations for government policies and consider integrating factors like pedestrian and cyclist safety, public transport prioritization, willingness-to-pay assessments, and equity considerations within the ATSC system design. This holistic approach, encompassing cost-benefit considerations, will form the foundation for more balanced and comprehensive traffic management solutions.

**Adaptive Traffic Signal's Safety and Efficiency Improvement by Multi-Objective Deep Reinforcement Learning Approach**

**Adaptive Traffic Signal's Safety and Efficiency Improvement by Multi-Objective Deep Reinforcement Learning Approach**